\documentclass[lettersize,journal]{IEEEtran}
\usepackage{amsmath,amsfonts}
\usepackage{algorithm}
\usepackage{array}
\usepackage[caption=false,font=normalsize,labelfont=sf,textfont=sf]{subfig}
\usepackage{textcomp}
\usepackage{stfloats}
\usepackage{url}
\usepackage{verbatim}
\usepackage{graphicx}
\usepackage{cite}
\usepackage{wrapfig}
\usepackage{color, colortbl}
\usepackage{makecell}
\usepackage{pifont}
\usepackage[most]{tcolorbox}
\newcommand{\cmark}{\ding{51}}
\newcommand{\xmark}{\ding{55}}
\definecolor{lightgray}{gray}{0.9}
\definecolor{green}{rgb}{0.9, 0.95, 0.97}
\newcolumntype{P}[1]{>{\centering\arraybackslash}p{#1}}
\usepackage{booktabs}

\usepackage{algorithm}
\usepackage{algpseudocode}
\usepackage{multirow}

\begin{document}

\title{DIP-R1: Deep Inspection and Perception with RL \\ Looking Through and Understanding Complex Scenes}

\author{Sungjune Park$^{\dagger}$, Hyunjun Kim$^{\dagger}$, Junho Kim, Seongho Kim, and Yong Man Ro*,~\IEEEmembership{Senior Member,~IEEE}
\thanks{S. Park, H. Kim, J. Kim, S. Kim, and Y. M. Ro are with Integrated Vision and Language Lab., School of Electrical Engineering, Korea Advanced Institute of Science and Technology (KAIST), 291 Daehak-ro, Yuseong-gu, Daejeon, 34141, Republic of Korea (E-mail: sungjune-p@kaist.ac.kr; kimhj709@kaist.ac.kr; arkimjh@kaist.ac.kr; taho43@kaist.ac.kr; ymro@kaist.ac.kr). \\ $^{\dagger}$Both authors contributed equally to this manuscript. \\ *Corresponding author: Yong Man Ro (E-mail: ymro@kaist.ac.kr).}}

\maketitle

\begin{abstract}
    Multimodal Large Language Models (MLLMs) have demonstrated significant visual understanding capabilities, yet their fine-grained visual perception in complex real-world scenarios, such as densely crowded public areas, remains limited. Inspired by the recent success of reinforcement learning (RL) in both LLMs and MLLMs, in this paper, we explore how RL can enhance visual perception ability of MLLMs. Then we develop a novel RL-based framework, \textbf{D}eep \textbf{I}nspection and \textbf{P}erception with \textbf{RL} (\textbf{DIP-R1}) designed to enhance the visual perception capabilities of MLLMs, by comprehending complex scenes and looking through visual instances closely. DIP-R1 guides MLLMs through detailed inspection of visual scene via three simply designed rule-based reward modelings. First, we adopt a standard reasoning reward encouraging the model to include three-step reasoning process: 1) comprehending entire visual scene, 2) observing for looking through interested but ambiguous regions, and 3) decision-making for predicting answer. Second, a variance-guided looking reward is designed to encourage MLLM to examine uncertain regions during the second observation process, guiding it to inspect ambiguous areas and thereby mitigate perceptual uncertainty. This reward further promotes variance-driven visual exploration, enabling the MLLM to reason about region-level uncertainty and explicitly indicate interpretable uncertain regions. Third, we model a weighted precision-recall accuracy reward enhancing accurate decision-making. We explore its effectiveness across diverse fine-grained object detection data consisting of challenging real-world environments, such as densely crowded scenes. Built upon existing MLLMs, DIP-R1 achieves consistent and significant improvement across various in-domain and out-of-domain scenarios. It also outperforms various existing baseline models and supervised fine-tuning methods. Our findings highlight the substantial potential of integrating RL into MLLMs for enhancing capabilities in complex real-world perception tasks.
\end{abstract}

\section{Introduction}
\begin{figure*}[t]
    \centering
    \includegraphics[width=1.0\textwidth]{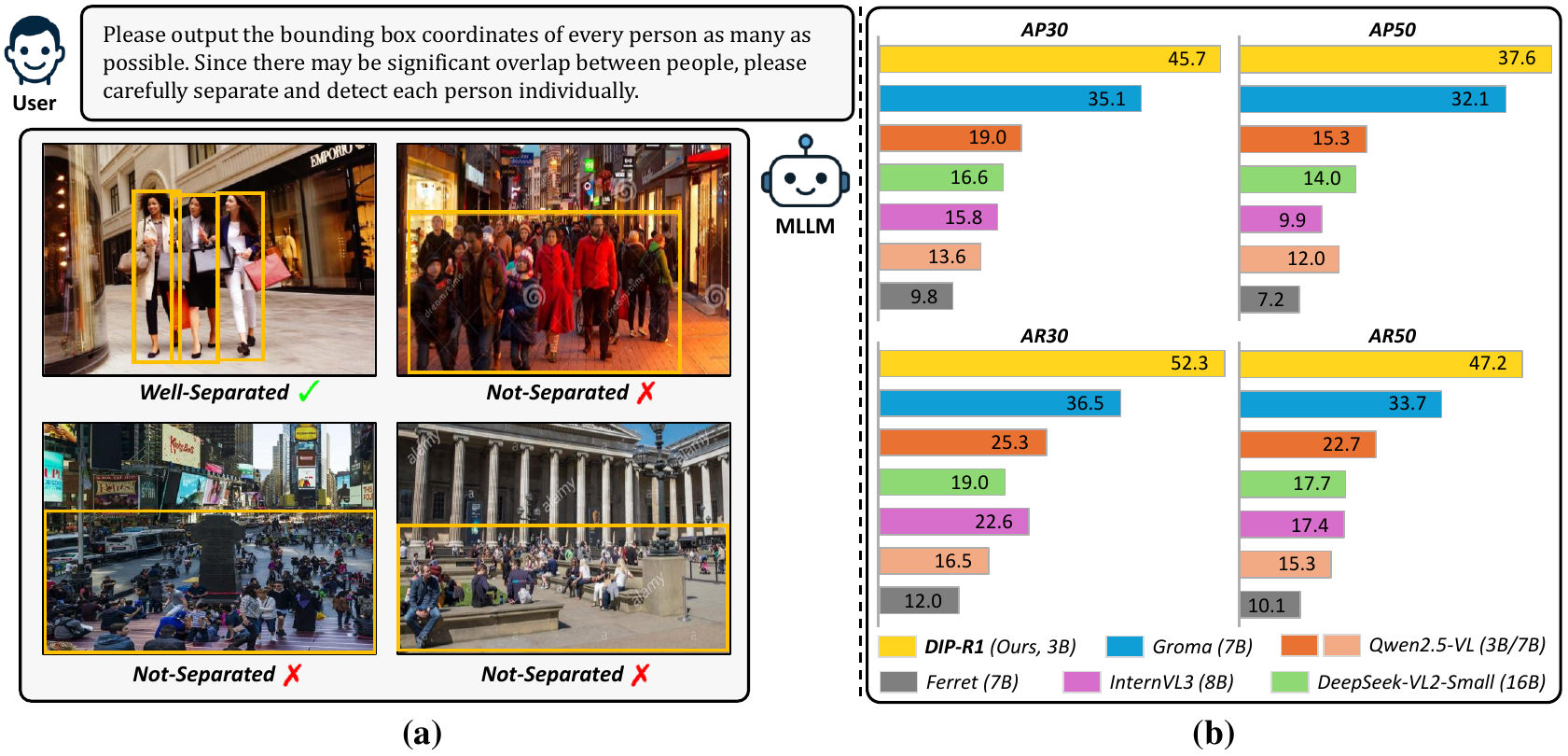}
    \vspace{-0.5cm}
	\caption{(a) shows public scene examples. While Qwen2.5-VL (3B) \cite{qwen2.5-vl} properly separate each person within the image including relatively large and less overlapped people. However, It fails to separate individuals and consider them as a single entity in the complex scenes. (b) shows that the proposed DIP-R1 obtains large improvements compared to existing methods.}
    \vspace{-0.3cm}
	\label{fig1}
\end{figure*}

\IEEEPARstart{T}{hese} days, multimodal large language models (MLLMs) have achieved remarkable progress \cite{cogvlm, mplug, minigpt, flamingo, gpt4v, gemini, internvl, tip1, tip2}, accelerated by advances in large language models (LLMs) \cite{vicuna, llama, llama2, llama3, phi3, qwen, alpaca, internlm, gpt4}, large-scale vision-language pretraining \cite{clip, blip2, tip3}, instruction tuning \cite{instructblip, llava}, and multimodal reasoning \cite{cot, deepseek-vl2, vision-r1, tip4, tip6}. These developments have significantly improved visual understanding capabilities of MLLMs, enabling noticeable performance across a wide range of visual understanding tasks, such as visual question answering (VQA), referring expression comprehension (REC), and image-text retrieval \cite{qwen2-vl, visionllm, tip5}. More recently, following the success of Reinforcement Learning (RL) in improving the reasoning abilities of LLMs, several studies have explored its application to visual domain tasks, demonstrating its effectiveness to enhance visual understanding and generalization with a small amount of data \cite{vision-r1, vlm-r1, r1-vl, visual-rft, cosmos-r1, ui-r1}. However, despite these advancements, MLLMs still suffer from perceiving complex real-world scenes, such as public places, where MLLMs can help human decision-making for public safety \cite{decision}. Such public scenes usually involve densely populated environments where many instances appear and overlap. For example, Fig.~\ref{fig1}(a) shows example public scenes and the visual perception (i.e., person detection) results of Qwen2.5-VL-3B-Instruct \cite{qwen2.5-vl}. As shown in the figure, while it properly perceives each person separately in the relatively easy environment (described in the first top-left example), it fails to capture each individual in crowded scenes, considering them as a single group. To mitigate this problem, MLLMs need to enhance their fine-grained visual perception ability by separating, recognizing, and localizing each instance properly, instead of regarding them as a single entity.

In this paper, we explore the way to enhance MLLMs’ fine-grained visual perception capability, based on recent findings and open-sources about RL \cite{deepseek-r1, vision-r1, r1-vl, vlm-r1}. By utilizing the advantages of RL, we encourage MLLMs to perceive visual scene semantics and each object separately in challenging real-world environments. Specifically, we aim to explicitly guide MLLM to articulate its step-by-step processes. To this end, we employ three simple rule-based reward modelings. First, we employ a standard \textit{format reward} for MLLM to include three step-by-step processes for each \textit{reasoning}, \textit{observing}, and \textit{decision-making}. MLLM looks through scenes and explains visual semantics especially in the scene level, such as, scene places, instance arrangements, and so on. Second, we design a \textit{variance-guided looking reward} for the observing process. Here, the variance is estimated to represent how much MLLM is uncertain its response, especially about region prediction. Due to this variance-guided looking process, the observing process is reinforced to include uncertain regions (which have high variances). Therefore, it enables MLLM to densely inspect uncertain region candidates which are confusing for being recognized properly. In other words, this variance-driven visual exploration guides MLLM to reason over region-level uncertainty. Third, we incorporate a \textit{weighted precision-recall reward} for decision-making process to minimize missing instances, predict correct region candidates, and give accurate answer. By integrating these components, we develop \textbf{D}eep \textbf{I}nspection and \textbf{P}erception \textbf{RL} framework, named \textbf{DIP-R1}, which can look through and comprehend visual scenes within challenging real-world environments.

We investigate the visual understanding ability in complex scenes with MLLMs' object detection performance in wild scenarios including densely crowded scenes, driving street areas, and aerial safety monitoring views, and so on. Extensive experimental results demonstrate that the proposed DIP-R1 framework is significantly effective in boosting MLLMs' visual perception capability in such challenging environments. DIP-R1 framework achieves noticeable improvements on four real-world object detection datasets, CrowdHuman \cite{crowdhuman}, CityPersons \cite{cityperson}, WiderPedestrian \cite{widerped}, and UAVDT \cite{uavdt}. Fig.~\ref{fig1}(b) shows detection performance on CrowdHuman and comparison with other baseline models and supervised fine-tuning (SFT) method on same Qwen2.5-VL baseline. In this work, we explore how RL is effective in improving visual perception capability of MLLMs in complex scenes. The contribution of our work can be summarized as follows:
\begin{itemize}
    \item{We investigate the limitation of current MLLMs suffering from difficulties in distinguishing and recognizing instances in challenging real-world environments, such as densely crowded areas, so that we develop a RL-based framework, DIP-R1.}
    \item{We incorporate three simple yet effective rule-based reward modelings: 1) a format reward to enforce reasoning, observing, and decision-making: three-step reasoning process, 2) a variance-guided looking reward to promote inspection of uncertain regions, and 3) a weighted precision-recall reward to perform robust instance-level recognition by securing balanced precision and recall.}
    \item{The explicit variance-driven visual exploration is incorporated in the reasoning process of MLLM, so that it enables MLLM to inspect uncertain regions and reason over region-level uncertainty by itself.}
    \item{We demonstrate the effectiveness of DIP-R1 through comprehensive experiments on four challenging real-world detection benchmarks, achieving large performance gain over existing baseline models including SFT.}
\end{itemize}

\section{Related Work}
\subsection{Multimodal Large Language Models (MLLMs)}
Thanks to recent advancements in LLMs \cite{vicuna, llama, llama2, llama3, phi3, qwen, alpaca, internlm, gpt4} and open sourcing of large-scale vision and language instruction datasets \cite{sharegpt4v, minigpt-v2, ferret, internlm-xcomposer, gpt4roi, glamm}, MLLMs have shown rapid progress, demonstrating remarkable visual understanding capabilities. Given user instructions, MLLMs mainly aim to understand and interpret visual semantics of a given image (or video). Most existing methods are designed to generate text description or select answers from a limited set of choices \cite{llavanext, llava-uhd, llava-uhd-v2, llava-hr, mg-llava}. To enhance visual understanding, LLaVA-Next \cite{llavanext}, MiniGemini \cite{mini-gemini}, and InternLM-XComposer2 \cite{internlm-xcomposer2} divide the input image into smaller sub-images. Furthermore, CoLLaVO \cite{collavo}, MoAI\cite{moai}, and Groma \cite{groma} deploy additional computer vision models, such as segmentation and region proposal networks, to improve their perception ability. Many works are built upon open-source frameworks, for example, LLaVA \cite{llava} and InternVL \cite{internvl}. Qwen2.5-VL \cite{qwen2.5-vl}, which is trained on diverse input modalities and at various scales, has demonstrated robust capabilities in accurately localizing objects using bounding boxes. In this paper, we adopt Qwen2.5-VL \cite{qwen2.5-vl} to build our DIP-R1 framework which effectively mitigates such a limitation within challenging real-world scenarios, even without using additional vision module.

\subsection{Reinforcement Learning (RL)}
RL has been considered a strong machine learning method, enabling agents to learn how to interact with environments and make decisions based on predefined reward policy \cite{rl}. More recently, the emergence of LLMs and powerful open-sourcing foundations, such as OpenAI-o1 \cite{openai-o1} and VLM-R1 \cite{vlm-r1}, has accelerated significant progress in applying RL for aligning language models with human preferences. Pioneering approaches include Reinforcement Learning with Human Feedback (RLHF) \cite{rlhf, instructgpt}, Proximal Policy Optimization (PPO) \cite{ppo}, Direct Preference Optimization (DPO) \cite{dpo}, and Kahneman-Tversky Optimization (KTO) \cite{kto}. Recent works (e.g., ReST-MCTS\* \cite{rest-mcts} and DeepSeek-R1 \cite{deepseek-r1}) have demonstrated that simple rule-based or outcome-level reward signals can effectively improve the reasoning abilities of LLMs. In particular, DeepSeek-R1 \cite{deepseek-r1} employ Group Relative Policy Optimization (GRPO) \cite{deepseekmath}, which aggregates and compares group-level and token-level preferences. Following these advances, RL has been increasingly extended to MLLMs. While early efforts mainly focused on mitigating hallucinations and improving alignment through preference feedback, more recent works enhance multimodal reasoning and understanding via RL-based fine-tuning. For example, R1-OneVision \cite{r1-onevision} aims to bridge the gap between visual perception and reasoning by integrating Chain-of-Thought (CoT) approach with GRPO. R1-VL \cite{r1-vl} presents StepGRPO, which improves structured reasoning by providing step-wise accuracy and validity rewards. Visual-RFT \cite{visual-rft} moves beyond mathematical reasoning to visual understanding of MLLMs, leveraging a rule-based Intersection-of-Union (IoU) reward function with GRPO. VLM-R1 \cite{vlm-r1} also incorporates GRPO into visual domain to enhance MLLMs' generalization and understanding of open-world instances.

Building on this line of research, our work explores how RL can further improve the visual perception abilities of MLLMs, particularly in complex and challenging environments. Different from existing RL-based works, our framework incorporate additional reasoning process (i.e., observing process)--a variance-driven visual exploration helping MLLM inspect and reason over uncertain regions by itself.

\section{Preliminary}
\subsection{Group Relative Policy Optimization (GRPO)}
GRPO is a reinforcement learning algorithm frequently used for post-training of LLMs, particularly to enhance their reasoning capabilities \cite{deepseek-r1, deepseekmath}. GRPO operates by sampling group responses and updating the policy model based on their relative quality, while constraining deviations from a reference model for the stability. Specifically, given an input query, GRPO samples a group of responses $\boldsymbol{G}=\{o_1, o_2, \cdots, o_G\}$. Each response is evaluated by predefined rule-based reward functions, and then corresponding reward values are obtained by $\boldsymbol{R}=\{r_1, r_2, \cdots, r_G\}$. Then the relative advantages are computed by $\hat{A}_{i,t} = (r_i - \bar{r})/\text{std}(r)$, where $\bar{r}$ and $\text{std}(r)$ are mean and standard deviation of the group rewards, respectively. This relative advantage $\hat{A}_{i,t}$ is used as the normalized signal to encourage the model to act in a better way than the group average. With $\hat{A}_{i,t}$, the objective of GRPO is formulated as follows:
{\small
    \begin{equation}
        \begin{aligned}
            &\mathcal{L}_{\text{GRPO}}(\theta) = 
            - \frac{1}{\sum_{i=1}^{G} |o_i|} \sum_{i=1}^{G} \sum_{t=1}^{|o_i|}
            \Bigg[
            \min\left(
            \frac{\pi_\theta(o_{i,t} \mid q, o_{i,<t})}{\pi_{\theta_{\text{old}}}(o_{i,t} \mid q, o_{i,<t})} \hat{A}_{i,t}, \right. \\ &
            \left. \text{clip} \left(
            \frac{\pi_\theta(o_{i,t} \mid q, o_{i,<t})}{\pi_{\theta_{\text{old}}}(o_{i,t} \mid q, o_{i,<t})},
            1 - \epsilon, 1 + \epsilon
            \right) \hat{A}_{i,t}
            \right)
            - \beta \, \mathbb{D}_{KL}[\pi_\theta \parallel \pi_{\text{ref}}]
            \Bigg],
        \end{aligned}
    \end{equation}
}
where $\pi_{\theta_{\text{old}}}$, $\pi_{\theta}$, and $\pi_{\theta_{\text{ref}}}$ denote the old policy, current policy, and the reference models, respectively. $\epsilon$ and $\beta$ are hyper-parameters to control the clipping range and the penalty for divergence from the reference model. An important key feature of GRPO is its iterative sampling of group responses and relative rewards, which enables stable policy learning. In our work, we adopt GRPO to update our framework, \textit{DIP-R1}, for enhancing MLLMs' visual perception capabilities.

\subsection{Uncertainty Estimation in Neural Networks}
In deep neural networks (DNNs), various studies have investigated how to measure the variance of model predictions \cite{uncer1, uncer2, uncer3}. For example, \textit{epistemic uncertainty} has been introduced to estimate how much the deep learning model is uncertain about its prediction because of limited training data or imperfect learning \cite{uncer1}. This kind of uncertainty, which is also referred to as \textit{predictive uncertainty}, estimates the uncertainty in the model parameters \cite{uncer2}. A common approach to measure this uncertainty is by Monte-Carlo (MC) dropout approximation \cite{uncer3}, where the model's outputs are sampled multiple times. Specifically, while sampling $T$ predictions, the predictive mean is measured by:
\begin{equation}
    \mathbb{E}[y] \approx \frac{1}{T} \sum_{t=1}^{T} \hat{y}_t,
\end{equation}
where $\hat{y}_t$ is the model's prediction probability at the $t$-th sampling. The corresponding predictive variance--epistemic uncertainty--is calculated by:
\begin{equation}
    \text{Var}[y] \approx \frac{1}{T} \sum_{t=1}^{T} \hat{y}_t^2 - \left( \mathbb{E}[y] \right)^2.
\end{equation}
Given that GRPO is also a sampling-based algorithm which generates $G$ group responses for policy optimization, a question naturally arises: \textit{``Can we also estimate the model's variance during GRPO?''}. Inspired by this conceptual similarity, in this work, we explore whether MLLMs can assess the variance across group responses during GRPO, and whether this variance can be effectively used as a reward signal to improve visual understanding.

\section{Proposed Framework: DIP-R1}
\begin{figure*}[t]
    \centering
    \includegraphics[width=1.0\textwidth]{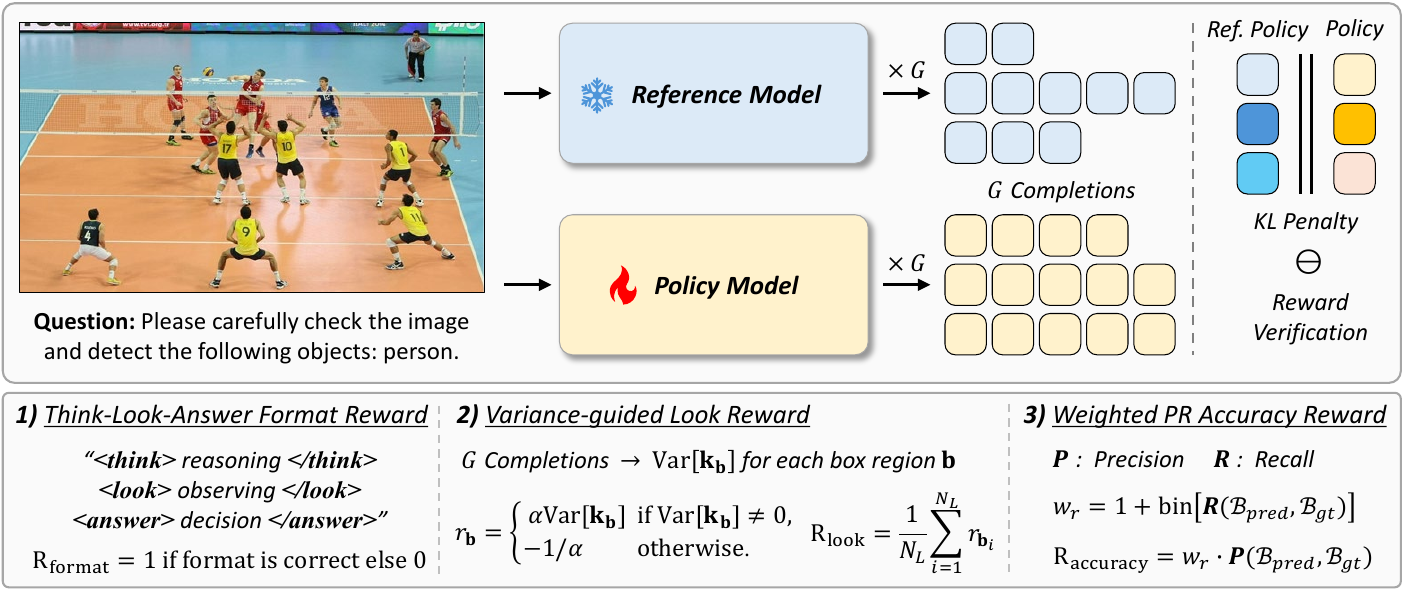}
	\caption{\textbf{Overall framework of the proposed DIP-R1}. Given an input sample, both the reference and policy models generate responses for KL regularization. DIP-R1 computes rewards based on three reward functions: 1) Think-Look-Answer format reward to include \texttt{<think>} reasoning \texttt{</think><look>} observing \texttt{</look><answer>} decision \texttt{</answer>}, 2) Variance-guided look reward to examine uncertain regions in observing process, and 3) Weighted precision-recall reward to obtain accurate answer by considering both precision and recall.}
	\label{fig2}
\end{figure*}

In this section, we describe the details about our proposed framework, \textbf{D}eep \textbf{I}nspection and \textbf{P}erception \textbf{RL} framework (\textbf{DIP-R1}).
Fig.~\ref{fig2} illustrates the overall architecture, which incorporates GRPO with three simple rule-based reward modelings: 1) \textit{think-look-answer format reward} which encourages step-by-step reasoning, observing, and decision-making processes, 2) \textit{variance-guided look reward} which helps look through uncertain regions, and 3) \textit{Weighted Precision-Recall Accuracy Reward} which considers both precision and recall to predict accurate answer. Our default instruction template and example output format are shown below. The more details about our reward modelings are elaborated in the following subsections.

\begin{figure*}[t]
    \begin{tcolorbox}[colback=gray!10, colframe=gray!70!black, title=Default Instruction Template, sharp corners=south]
    First, \textbf{think about the reasoning process}, and then \textbf{look carefully with observing process} in the mind and then provides the user with the answer. \textbf{The observing process must include the coordinates of the any ambiguous regions where objects overlap or are partially occluded.} The reasoning process and observing process and answer are enclosed within \texttt{<think> </think>} and \texttt{<look> </look>} and \texttt{<answer> </answer>} tags, respectively, i.e., \texttt{<think>} reasoning process here \texttt{</think><look>} observing process here \texttt{</look><answer>} answer here \texttt{</answer>}. \{\textit{Question}\}. Output the bbox coordinates of detected objects in \texttt{<answer></answer>}. The bbox coordinates in Markdown format should be: \textbackslash n`` `json\textbackslash n[\{\{``bbox\_2d'':[x1, y1, x2, y2], ``label'': ``object name''\}\}]\textbackslash n'' '\textbackslash n If not targets are detected in the image, simply respond with ``None''.
    \end{tcolorbox}
\end{figure*}

\begin{figure*}[t]
    \begin{tcolorbox}[colback=gray!10, colframe=gray!70!black, title=Example Output Format, sharp corners=south]
    \texttt{<think>} First, I'll inspect the image to identify each person present. The group consists of around 20 individuals standing closely and posing for a photo. Since the task is to identify individual persons in the crowd, $\cdots$ \texttt{</think>}
    
    \texttt{<look>} [\{``bbox\_2d'': [124, 423, 183, 693], ``label'': ``person''\}, \{``bbox\_2d'': $\cdots$] \texttt{</look>}
    
    \texttt{<answer>} [\{``bbox\_2d'': [258, 451, 343, 845], ``label'': ``person''\}, \{``bbox\_2d'': $\cdots$] \texttt{</answer>}
    \end{tcolorbox}
\end{figure*}

\subsection{Think-Look-Answer Format Reward}
We design the think-look-answer format reward to encourage MLLM to infer the answer through three step-by-step processes: scene-level reasoning, region-level observing, and decision-making. Based on the conventional format reward for \texttt{<think> ... </think><answer> ... </answer>}, we add \texttt{<look> ... </look>} between them. Therefore, MLLM learns to generate responses following three step-by-step processes: reasoning, observing, and decision-making. The Think-Look-Answer reward assigns 1 if the response includes all three components in the correct order, and 0 otherwise:
\begin{equation}
    \mathrm{R}_{\text{format}} = \begin{cases} 1, & \text{if the response follows the correct format,} \\ 0, & \text{otherwise.} \end{cases}
\end{equation}

Within this structure, the reasoning process captures scene-level descriptions, such as place information, spatial layouts and relationships between instances. The observing process inspects uncertain regions to enhance fine-grained perception, and the details are described in the next subsection.

\subsection{Variance-guided Look Reward}
In challenging, complex environments, we expect that MLLM would be easily confused to recognize instances properly. Therefore, we aim to estimate region variance and help MLLM recognize such ambiguous instances. Therefore, the variance-guided look reward is designed to encourage MLLM to include and examine ambiguous or confusing regions in the observing process. But how can we identify such uncertain regions? Inspired by sampling-based uncertainty estimation methods \cite{uncer2, uncer3, uncer4}, we recognize that GRPO is also a sampling-based algorithm generating $G$ different responses. Specifically, we probe MLLM's output within \texttt{<answer></answer>} to identify regions where MLLM shows inconsistent predictions across samples--these are treated as uncertain regions. In other words, a region is considered uncertain, if MLLM inconsistently generates it across $G$ responses. Let $\mathbf{B} =\{\mathcal{B}_1, \mathcal{B}_2, \cdots, \mathcal{B}_G\}$ denote the sets of region boxes predicted during $G$ generations. The frequency (or occurrence probability) of a specific region box $\mathbf{b}$ can be computed as follows:
\begin{equation}
    f_{\mathbf{b}} = \frac{1}{G} \sum_{i=1}^{G} \boldsymbol{1}\left[ \exists \tilde{\mathbf{b}} \in \mathcal{B}_i \ \text{s.t.} \ \text{IoU}(\mathbf{b}, \tilde{\mathbf{b}}) \geq \tau \right],
\end{equation}
where IoU($\cdot$) represents the Intersection-of-Union (IoU) function, and $\tau$ is the IoU threshold to determine whether $\mathbf{b}$ and $\tilde{\mathbf{b}}$ represent the same region. This equation counts how many times a region equivalent to $\mathbf{b}$ appears across the $G$ responses. Assuming that the generation process for each region box $\mathbf{b}$ follows an independent Bernoulli trial with occurrence probability $f_{\mathbf{b}}$, the total number of occurrences across $G$ samples follows a Binomial distribution $\mathbf{k_b} \sim \text{Binomial}(G, f_{\mathbf{b}})$. Accordingly, the variance of $\mathbf{k}_{\mathbf{b}}$, which reflects the uncertainty of region $\mathbf{b}$, is obtained as follows:
\begin{equation}
    \text{Var}[\mathbf{k_b}] =G f_{\mathbf{b}} (1 - f_{\mathbf{b}}) \approx f_{\mathbf{b}} (1 - f_{\mathbf{b}}),
\end{equation}
where we omit the constant $G$. The variance will be maximum when $f_{\mathbf{b}}=0.5$ (i.e., MLLM predicts the region $\mathbf{b}$ half of G times), and be zero when $f_{\mathbf{b}} = 0$ or $1$ (i.e., MLLM is fully certain). we employ this variance to guide MLLM to include uncertain regions within \texttt{<look></look>} tags, and the variance-guided look reward is formulated as follows:
\begin{equation}
    r_{\mathbf{b}} =
    \begin{cases}
        \alpha\text{Var}[\mathbf{k_b}], & \text{if } \text{Var}[\mathbf{b}] \neq 0, \\
        -1/\alpha, & \text{otherwise},
    \end{cases}
\end{equation}
\begin{equation}
    \mathrm{R}_{\text{look}} = \frac{1}{N_L}\sum_{i=1}^{N_L} r_{\mathbf{b}_i},
\end{equation}
where $N_L$ is the number of region boxes observed in the observing process, and $\alpha$ is a normalization factor (dependent on $G$) to scale the reward to [0, 1] when $\text{Var}[\mathbf{k_b}] \neq 0$. Otherwise, a small penalty is applied when certain regions are included (i.e., $\text{Var}[\mathbf{k_b}]=0$). In summary, the more uncertain regions are involved in the observing process, the higher reward is be acquired. As a result, due to this variance-guided look reward, the observing process learns to inspect ambiguous regions in complex scenes that MLLM is confused. Algorithm~\ref{alg1} is the pseudo-code elaborating how to compute the variance-guided look reward in details.

\begin{algorithm}[t]
    \caption{The way to Compute Variance-guided Look Reward}
\begin{algorithmic}[1]
    \Require Number of generations $G$, Box sets in each response $\{\mathcal{B}_1, \dots, \mathcal{B}_G\}$, IoU threshold $\tau$, Normalization parameter $\alpha$, Box sets observed in \texttt{<look>} tags $\{\mathcal{B}^l_1, \dots, \mathcal{B}^l_G\}$
    \State Initialize set of all boxes $\mathcal{B} \gets \bigcup_{i=1}^{G} \mathcal{B}_i$ \Comment{All predicted boxes from $G$ responses}
    \For{each box $\mathbf{b} \in \mathcal{B}$}
        \State $f_b \gets 0$
        \For{$i=1, \dots, G$}
            \If{$\exists \tilde{\mathbf{b}} \in \mathcal{B}_i \text{ s.t. } \text{IoU}(\mathbf{b}, \tilde{\mathbf{b}}) \geq \tau$}
                \State $f_b \gets f_b + 1$ \Comment{Count the frequency of same box across generations}
                \State $continue$
            \EndIf
        \EndFor
        \State $f_b \gets f_b / G$
        \State $\text{Var}[\mathbf{k}_{\mathbf{b}}] \gets f_b(1-f_b)$ \Comment{Compute variance}
        \If{$\text{Var}[\mathbf{k}_{\mathbf{b}}] > 0$}
            \State $r_{\mathbf{b}} \gets \alpha \cdot \text{Var}[\mathbf{k}_{\mathbf{b}}]$ \Comment{Compute box reward based on the variance}
        \Else
            \State $r_{\mathbf{b}} \gets -1/\alpha$ \Comment{Penalty for certain boxes}
        \EndIf
    \EndFor
    \State $\mathrm{R}_{\text{look}}=[\;]$
    \For{each box set $\mathcal{B}^l_i \subset \{\mathcal{B}^l_1, \dots, \mathcal{B}^l_G\}$}
        \State Let $\mathcal{B}^l_i = \{\mathbf{b}_1, \dots, \mathbf{b}_{N_L}\}$
        \State $\mathrm{R}_i \gets \frac{1}{N_L} \sum_{j=1}^{N_L} r_{\mathbf{b}_j}$ \Comment{Compute Look Reward for each response}
        \State $\mathrm{R}_{\text{look}}\text{.append}(\mathrm{R}^i)$
    \EndFor
    \State \Return $\mathrm{R}_{\text{look}}$
\end{algorithmic}
\label{alg1}
\end{algorithm}

\subsection{Precision-Recall Accuracy Reward}
\label{pr-reward}
In the decision making process, the model predicts answer between \texttt{<answer></answer>}. In our work, the answer is bounding box candidates and their labels relevant to the user query. Therefore, to perform accurate decision making process in \texttt{<answer></answer>}, we design weighted precision-recall accuracy reward. While the model generates a list of bounding box regions $\mathcal{B}_{ans}=\{\mathbf{b}_1, \mathbf{b}_2, \cdots, \mathbf{b}_A\}$, it calculates the precision and recall by comparing them with ground-truth bounding box coordinates $\mathcal{B}_{gt}$. We use IoU threshold 0.5 to determine whether each answer box is correct or not. Then we formulate weighted precision-recall accuracy reward as follows:
\begin{equation}
    \mathrm{R}_{\text{accuracy}} = w_{r} \cdot \boldsymbol{P}(\mathcal{B}_{pred}, \mathcal{B}_{gt}),
\end{equation}
\begin{equation}
    w_{r} = 1 + \text{bin}\left[\boldsymbol{R}(\mathcal{B}_{pred}, \mathcal{B}_{gt})\right],
\end{equation}
where $\boldsymbol{P}(\cdot)$ and $\boldsymbol{R}(\cdot)$ are the functions calculating precision and recall metrics, respectively. $\text{bin}\left[\boldsymbol{R}(\mathcal{B}_{pred}, \mathcal{B}_{gt})\right]$ discretizes a recall value into one of 100 equal intervals over the range [0, 1], so that $w_r$ denotes a weight which is linearly increasing depending on recall. Since a simple precision reward modeling considers precision only, so that it fails to decrease \textit{false negatives}. Different from the simple precision reward, this reward design takes into account recall value, so that it becomes large when both precision and recall are high. As a result, as the model's answer is more accurate (i.e., high precision) and covers as many as relevant regions (i.e., high recall), this reward modeling can assign large reward to the policy. Owing to this reward, our DIP-R1 framework generates more accurate answers even within crowded scenes where many objects appear and overlapped.
\begin{table*}[t!]
    \centering
    \caption{Performance comparison with baseline models on in-domain (ID) CrowdHuman \cite{crowdhuman} and out-of-domain (OOD) CityPersons \cite{cityperson}, WiderPedestrian \cite{widerped}, and UAVDT \cite{uavdt}. The best results are highlighted in \textbf{bold} and the runner-up results are \underline{underlined}.}
    \renewcommand{\arraystretch}{1.1}
    \renewcommand{\tabcolsep}{1.0mm}
    \resizebox{0.999\linewidth}{!}
    {
        \renewcommand{\arraystretch}{1.2}
        \begin{tabular}{l|P{1.2cm}P{1.2cm}P{1.2cm}P{1.2cm}|P{1.2cm}P{1.2cm}P{1.2cm}P{1.2cm}}
            \toprule[1.3pt]
            \multirow{2}{*}{\bf Model} & \bf AP30 & \bf AR30 & \bf AP50 & \bf AR50 & \bf AP30 & \bf AR30 & \bf AP50 & \bf AR50 \\
            \cmidrule{2-9}
             & \multicolumn{4}{c|}{\bf \footnotesize CrowdHuman (ID)} & \multicolumn{4}{c}{\bf \footnotesize CityPersons (OOD)} \\
            \midrule
            Ferret (7B) \cite{ferret}& 9.8 & 12.0 & 7.2 & 10.1 & 4.7 & 7.0 & 1.9 & 4.5 \\
            Qwen2.5-VL-Instruct (3B) \cite{qwen2.5-vl}& 13.6 & 16.5 & 12.0 & 15.3 & 9.4 & 14.1 & 6.5 & 11.7 \\
            InternVL3 (8B) \cite{internvl3}& 15.8 & 22.6 & 9.9 & 17.4 & 6.4 & 13.0 & 3.2 & 8.3 \\
            DeepSeek-VL2-Small (16B) \cite{deepseek-vl2}& 16.6 & 19.0 & 14.0 & 17.7 & 15.5 & \bf 30.8 & 10.3 & \underline{24.5} \\
            Qwen2.5-VL-Instruct (7B) \cite{qwen2.5-vl}& 19.0 & 25.3 & 15.3 & 22.7 & 15.5 & 26.3 & 9.9 & 20.3 \\
            Groma (7B) \cite{groma}& 35.1 & 36.5 & 32.1 & 33.7 & \underline{19.3} & 20.2 & \underline{15.1} & 16.2 \\
            \midrule
            \rowcolor{lightgray}
            \bf DIP-R1 (Qwen2.5-VL-Instruct, 3B) & \underline{41.8} & \underline{49.2} & \underline{35.5} & \underline{45.2} & 17.2 & 24.3 & 12.0 & 20.1 \\
            \rowcolor{lightgray}
            \bf DIP-R1 (Qwen2.5-VL-Instruct, 7B) & \bf 45.7 & \bf 52.3 & \bf 37.6 & \bf 47.2 & \bf 24.5 & \underline{30.7} & \bf 16.6 & \bf 24.6 \\ 
            \midrule \midrule
            \bf Model & \multicolumn{4}{c|}{\bf \footnotesize WiderPedestrian (OOD)} & \multicolumn{4}{c}{\bf \footnotesize UAVDT (OOD)} \\
            \midrule
            Ferret (7B) \cite{ferret}& 6.9 & 16.0 & 2.5 & 9.7 & 5.4 & 8.5 & 2.2 & 4.3 \\
            Qwen2.5-VL-Instruct (3B) \cite{qwen2.5-vl}& 14.1 & 30.8 & 11.3 & 27.5 & 9.0 & 20.1 & 7.9 & 18.6 \\
            InternVL3 (8B) \cite{internvl3}& 14.0 & 36.8 & 7.6 & 26.1 & 2.3 & 11.1 & 1.5 & 6.0 \\    
            DeepSeek-VL2-Small (16B) \cite{deepseek-vl2}& \underline{30.6} & \underline{51.8} & \underline{24.9} & 46.2 & 12.5 & 23.6 & 10.8 & 21.0 \\
            Qwen2.5-VL-Instruct (7B) \cite{qwen2.5-vl}& 19.8 & \underline{51.8} & 15.3 & 45.1 & 9.5 & 25.3 & 7.9 & 22.0 \\
            Groma (7B) \cite{groma}& 19.8 & 48.0 & 15.4 & 39.0 & 21.0 & 25.9 & \underline{19.3} & 23.9 \\
            \midrule
            \rowcolor{lightgray}
            \bf DIP-R1 (Qwen2.5-VL-Instruct, 3B) & 24.4 & 51.7 & 20.0 & \underline{46.7} & \underline{22.9} & \underline{40.0} & \bf 20.2 & \underline{36.9} \\
            \rowcolor{lightgray}
            \bf DIP-R1 (Qwen2.5-VL-Instruct, 7B) & \bf 32.8 & \bf 62.4 & \bf 26.2 & \bf 55.8 & \bf 24.3 & \bf 43.4 & 19.0 & \bf 37.3 \\
            \bottomrule[1.3pt]
        \end{tabular}
    }
    \label{tab1}
\end{table*}

\section{Experiment}
\subsection{Datasets}
To evaluate the effectiveness of our DIP-R1 framework in enhancing the visual perception capabilities of MLLMs in complex real-world scenes, we conduct experiments on four challenging real-world object detection datasets: CrowdHuman \cite{crowdhuman}, CityPersons \cite{cityperson}, WiderPedestrian \cite{widerped}, and UAVDT \cite{uavdt}. These datasets include a wide range of complex scenarios, such as densely crowded public areas, driving environments, and aerial safety monitoring views. CrowdHuman is one of the most widely-known person detection datasets. The images are collected by large-scale web crawling and include diverse, challenging environments such as both indoor and outdoor public places (e.g., sports facilities, plazas, and city streets). We use the training data of CrowdHuman to train our DIP-R1 framework, and its validation set is used for in-domain evaluation. CityPersons and WiderPedestrian are also well-known person detection datasets, covering driving street scenes and safety monitoring views. Their validation sets are used for the out-of-domain evaluation. We use the visible box annotations for the evaluation on CrowdHuman and CityPersons. In addition, we also evaluate our framework on the test set of UAVDT, which contains more challenging aerial-view scenarios. We randomly sample 500 images from the UAVDT test set for out-of-domain evaluation. Every baseline model and our DIP-R1 framework are evaluated on the same subset to ensure a fair comparison. After RL fine-tuning on CrowdHuman, we measure performance on both in-domain CrowdHuman and out-of-domain data, CityPersons, WiderPedestrian, and UAVDT. We report average precision and recall metrics based on IoU thresholds of 0.3 and 0.5 (AP30, AR30, AP50, and AR50).

\subsection{Implementation Details}
To construct our framework, we adopt Qwen2.5-VL-Instruct (3B and 7B) \cite{qwen2.5-vl} as our base model--one of the pioneering and open-sourcing MLLMs. We compare the performance of our DIP-R1 framework with several existing methods, including Ferret (7B) \cite{ferret}, InternVL3 (8B) \cite{internvl3}, DeepSeek-VL2-Small (16B) \cite{deepseek-vl2}, and Groma (7B) \cite{groma}. For training, we use 4 NVIDIA A100 80GB GPUs. The details of our GRPO training configuration are: number of generations $G$ is 8, batch size per device is 4, gradient accumulation steps are 4, the number of training epoch is 1, KL divergence coefficient $\beta$ is 0.04, maximum output length is 2048 tokens, and the learning rate is 1e-6. The rollout number is $4$ devices $\times\,4$ batch sizes/device $\times\,4$ accumulation steps $\times\,8$ generations $=\,512$. We report the best performance of our framework and SFT method within an epoch for the main comparison in TABLE~\ref{tab1}. DIP-R1 framework is built upon well-established RL foundation \cite{vlm-r1}.

\begin{figure*}[t!]
    \begin{minipage}[b]{0.99\linewidth}
        \centering
        \centerline{\includegraphics[width=\linewidth]{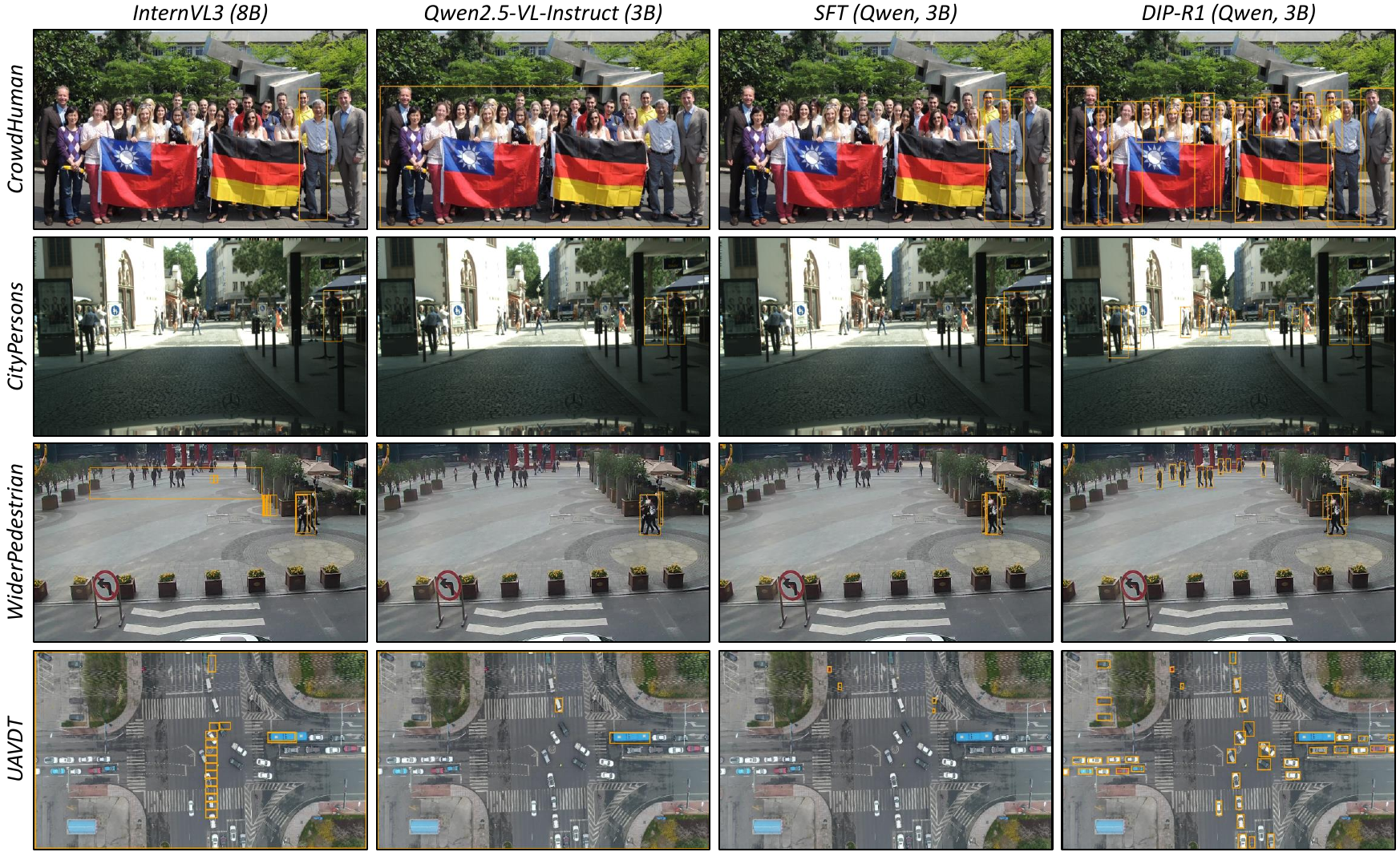}}
    \end{minipage}
	\caption{The visualization results of Qwen2.5-VL-Instruct (3B), SFT on Qwen2.5-VL-Instruct (3B), DeepSeek-VL2-Small (16B), and DIP-R1 (3B), along with ground-truth regions. As shown in the figure, the existing baseline models easily fail to perceive each individual and tend to miss or consider them as a single entity. On the other hand, our method separates and recognizes each of them properly.}
	\label{fig3}
\end{figure*}

\begin{figure*}[t!]
    \begin{minipage}[b]{0.99\linewidth}
        \centering
        \centerline{\includegraphics[width=\linewidth]{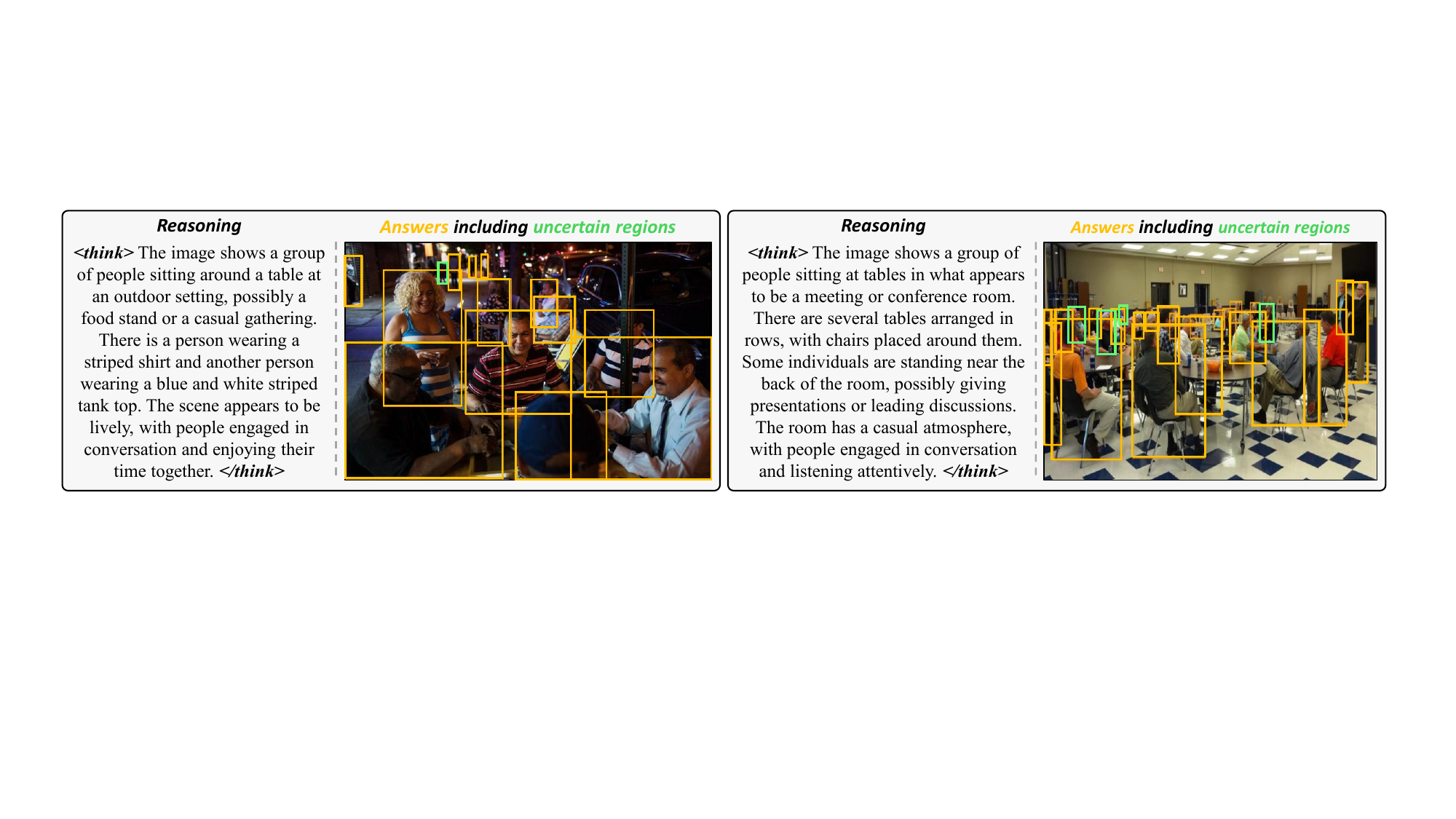}}
    \end{minipage}
	\caption{Output analysis of DIP-R1 framework, showing the reasoning description, the uncertain regions (\textit{green boxes}) captured in the observing process, and the prediction results (\textit{orange boxes}).}
	\label{fig4}
\end{figure*}

\subsection{Experimental Result}
\subsubsection{Comparison with Baselines}
In this section, we compare the performance of our DIP-R1 framework with various baseline models on four object detection datasets, including in-domain data: CrowdHuman \cite{crowdhuman} and three out-of-domain data: CityPersons \cite{cityperson}, WiderPedestrian \cite{widerped}, and UAVDT \cite{uavdt}. TABLE~\ref{tab1} shows the performance comparison with the other baseline models: Ferret (7B) \cite{ferret}, Qwen2.5-VL-Instruct (3B, 7B) \cite{qwen2.5-vl}, InternVL3 (8B) \cite{internvl3}, DeepSeek-VL2-Small (16B) \cite{deepseek-vl2}, and Groma (7B) \cite{groma}. On CrowdHuman, our framework shows remarkable performance gains over existing baselines, including DeepSeek-VL2-Small (16B) \cite{deepseek-vl2} and InternVL3 (8B) \cite{internvl3}, which are among the most recently released state-of-the-art MLLMs. Notably, even compared to Groma \cite{groma}--a grounding-specialized MLLM which adopts additional region proposal networks, our framework achieves superior performance. On the out-of-domain data, DIP-R1 framework consistently enhances visual perception capabilities across diverse, challenging environments.

\subsubsection{Qualitative Analysis} We further conduct a qualitative analysis to validate our framework built upon Qwen2.5-VL (7B). Fig.~\ref{fig3} shows the visualization results of baseline models and our framework along with ground-truth region boxes. Additionally, Fig.~\ref{fig4} illustrates the output of our framework, including the reasoning description, the uncertain regions (\textit{green boxes}) captured within the observing process, and the residual predictions (\textit{orange boxes}). As shown in the figure, the reasoning step provides a description of the overall scene context. We also analyze which regions are assigned to the uncertain regions. In the figure, blue boxes represent the uncertain regions, which are included in the observing process, and they tend to capture small or partially visible instances. More visualization results are included in Fig.~\ref{fig8}.

\begin{figure*}[t!]
    \begin{minipage}[b]{0.99\linewidth}
        \centering
        \centerline{\includegraphics[width=\linewidth]{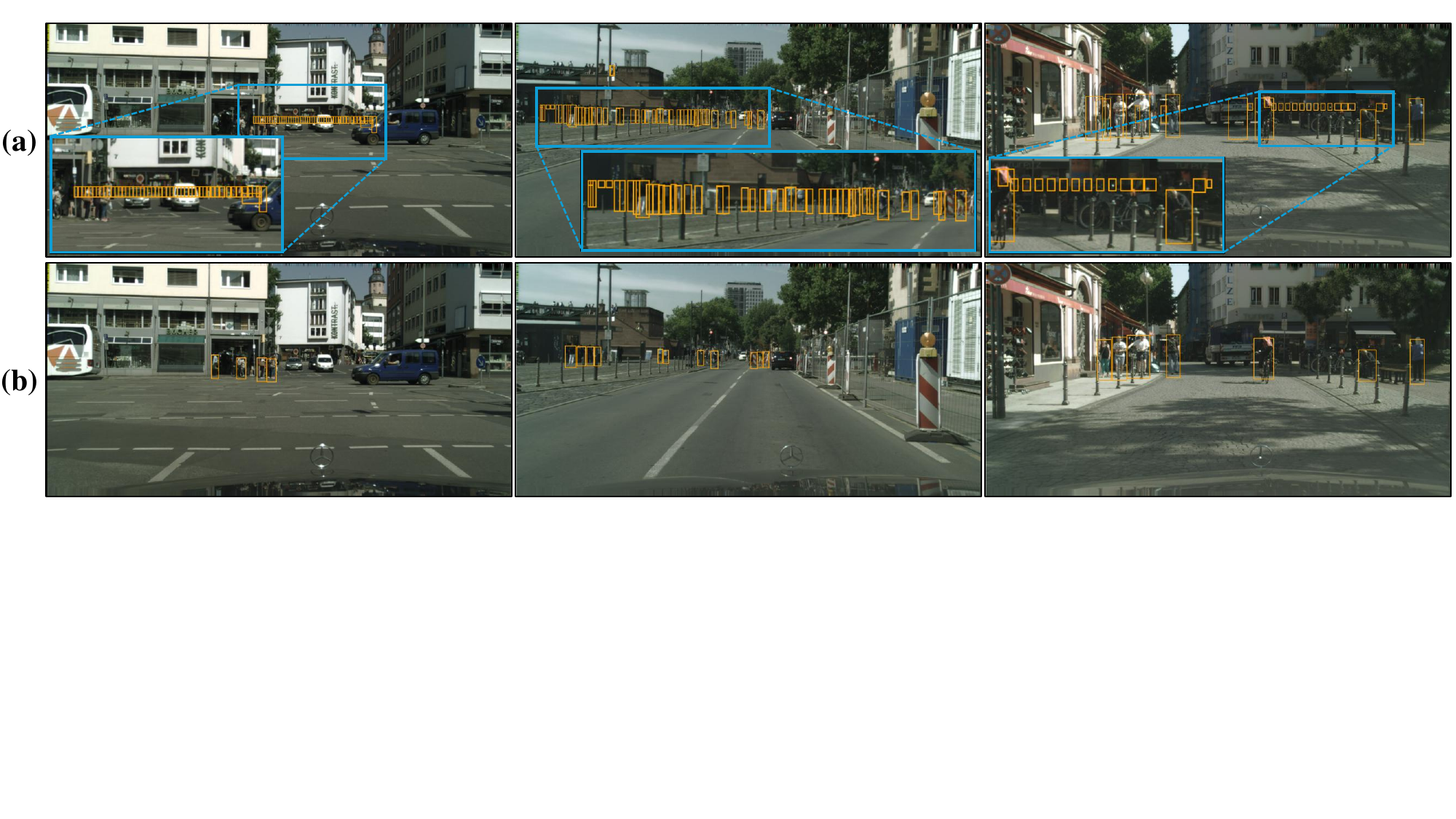}}
    \end{minipage}
	\caption{Qualitative comparative analysis between the SFT baseline and DIP-R1, where both are built upon Qwen2.5-VL-Instruct (3B). As described in (a), SFT usually exhibits noisy prediction results which may lead to high recall but low precision. On the other hand, as shown in (b), DIP-R1 performs more reliable prediction.}
	\label{fig5}
\end{figure*}

\begin{table*}[t!]
    \centering
    \caption{Performance comparison with SFT on in-domain (ID) CrowdHuman \cite{crowdhuman} and out-of-domain (OOD) CityPersons \cite{cityperson}, WiderPedestrian \cite{widerped}, and UAVDT \cite{uavdt}.}
    \renewcommand{\arraystretch}{1.1}
    \renewcommand{\tabcolsep}{5.0mm}
    \resizebox{0.999\linewidth}{!}
    {
        \renewcommand{\arraystretch}{1.3}
        \begin{tabular}{l|cc|cc|cc|cc}
            \toprule[1.3pt]
            \multicolumn{1}{c|}{} & \multicolumn{2}{c|}{\bf \footnotesize CrowdHuman (ID)} & \multicolumn{2}{c|}{\bf \footnotesize CityPersons (OOD)} & \multicolumn{2}{c|}{\bf \footnotesize WiderPedestrian (OOD)} & \multicolumn{2}{c}{\bf \footnotesize UAVDT (OOD)} \\
            \midrule
            \bf Method & \bf P@50 & \bf R@50 & \bf P@50 & \bf R@50 & \bf P@50 & \bf R@50 & \bf P@50 & \bf R@50 \\
            \midrule
            SFT & 33.3 & \bf 47.7 & 3.9 & 16.5 & 12.2 & 46.2 & 2.5 & 6.9   \\
            DIP-R1 & \bf 35.5 & 45.1 & \bf 12.0 & \bf 20.1 & \bf 20.0 & \bf 46.7 & \bf 21.3 & \bf 36.9  \\
            \bottomrule[1.3pt]
        \end{tabular}
    }
    \label{tab2}
\end{table*}

\begin{table}[t!]
    \centering
    \caption{The performance comparison (AP50) with SFT at each training step on CrowdHuman.}
    \renewcommand{\arraystretch}{1.1}
    \renewcommand{\tabcolsep}{3.0mm}
    \resizebox{0.999\linewidth}{!}
    {
        \begin{tabular}{l|cccccc}
            \toprule[1.3pt]
            \bf Method & \bf 100 & \bf 200 & \bf 300 & \bf 400 & \bf 500 & \bf 600 \\
            \midrule
            SFT & 15.9 & 16.3 & 21.3 & 25.5 & 29.1 & \bf 33.3 \\
            DIP-R1 & 17.4 & 25.9 & 30.8 & 29.8 & 33.9 & \bf 35.5 \\
            \midrule
            $\Delta$ & 1.5 & 9.6 & 9.5 & 4.3 & 4.8 & 2.2 \\
            \bottomrule[1.3pt]
        \end{tabular}
    }
    \label{tab3}
\end{table}

\subsubsection{Comparison with SFT}
In this section, we analyze the effectiveness of our framework by comparing it with SFT. The experiments are conducted with Qwen2.5-VL-Instruct (3B). First, TABLE~\ref{tab2} shows SFT performance on in-domain and out-of-domain data. As described in the table, our framework mostly shows superior perception and generalization performance. Notably, SFT methods exhibit particularly poor performance on out-of-domain classes and environments, indicating limited generalization capability beyond the training distribution. Second, TABLE~\ref{tab3} presents performance comparison with SFT on CrowdHuman at each training step. Our DIP-R1 framework consistently outperforms SFT across all training steps, demonstrating the effectiveness of RL in enhancing visual perception without relying on fully supervised training, using a small amount of data. Additionally, Fig.~\ref{fig5} presents a qualitative comparative analysis between the SFT baseline and our framework. As shown in the figure, SFT tends to produce noisy prediction results, which align with its quantitative result of high recall but low precision due to numerous false positives in TABLE~\ref{tab2}. This sequentially generative behavior may stem from the SFT's inherent tendency \cite{sft1, sft2} memorizing and imitating frequent patterns in the training data (i.e., crowded consecutive patterns). In contrast, our framework effectively mitigates such unintended behaviors.

\subsubsection{Ablation Study} We also conduct ablation study on CrowdHuman (CH, in-domain data) and CityPersons (CP, out-of-domain data) to validate the effectiveness of each reward modeling component. TABLE~\ref{tab4} shows the experimental result using Qwen2.5-VL-Instruct (3B). When there is the reasoning process only, it obtains 29.5 AP50 on CH and 8.4 AP50 on CP, respectively. Then when the observing process is integrated with the variance-guided look reward, it achieves 33.3 AP50 on CH with 6.3 gain and 11.3 AP50 on CP with 4.8 gain, showing the effectiveness on both in- and out-of-domain data. Lastly, as the weighted precision-recall accuracy reward is incorporated, it obtains 35.5 AP50 on CH and 12.0 AP50 on CP, respectively.

\begin{table}[t!]
    \centering
    \caption{Ablation study for each reward modeling with Qwen2.5-VL-Instruct (3B) on CrowdHuman (ID) and CityPersons (OOD). The evaluation metric is AP50.}
    \renewcommand{\arraystretch}{1.15}
    \renewcommand{\tabcolsep}{1.0mm}
    \resizebox{0.999\linewidth}{!}
    {
        \begin{tabular}{ccc|cc}
        \toprule[1.3pt]
        \bf Format & \bf Look & \bf PR & \bf CH & \bf CP \\ \midrule
        \texttt{<answer>} & \xmark & \xmark & 27.0 & 8.4 \\
        \texttt{<think>-<answer>} & \xmark & \xmark & 29.5 (+2.5) & 9.4 (+1.0) \\
        \texttt{<think>-<look>-<answer>} & \cmark & \xmark & 33.3 (+6.3) & 11.3 (+4.8) \\
        \texttt{<think>-<look>-<answer>} & \cmark & \cmark & \bf 35.5 (+8.3) & \bf 12.0 (+5.5) \\
        \bottomrule[1.3pt]
    \end{tabular}
    }
    \label{tab4}
\end{table}
\begin{figure}[t!]
    \begin{minipage}[b]{0.99\linewidth}
        \centering
        \centerline{\includegraphics[width=\linewidth]{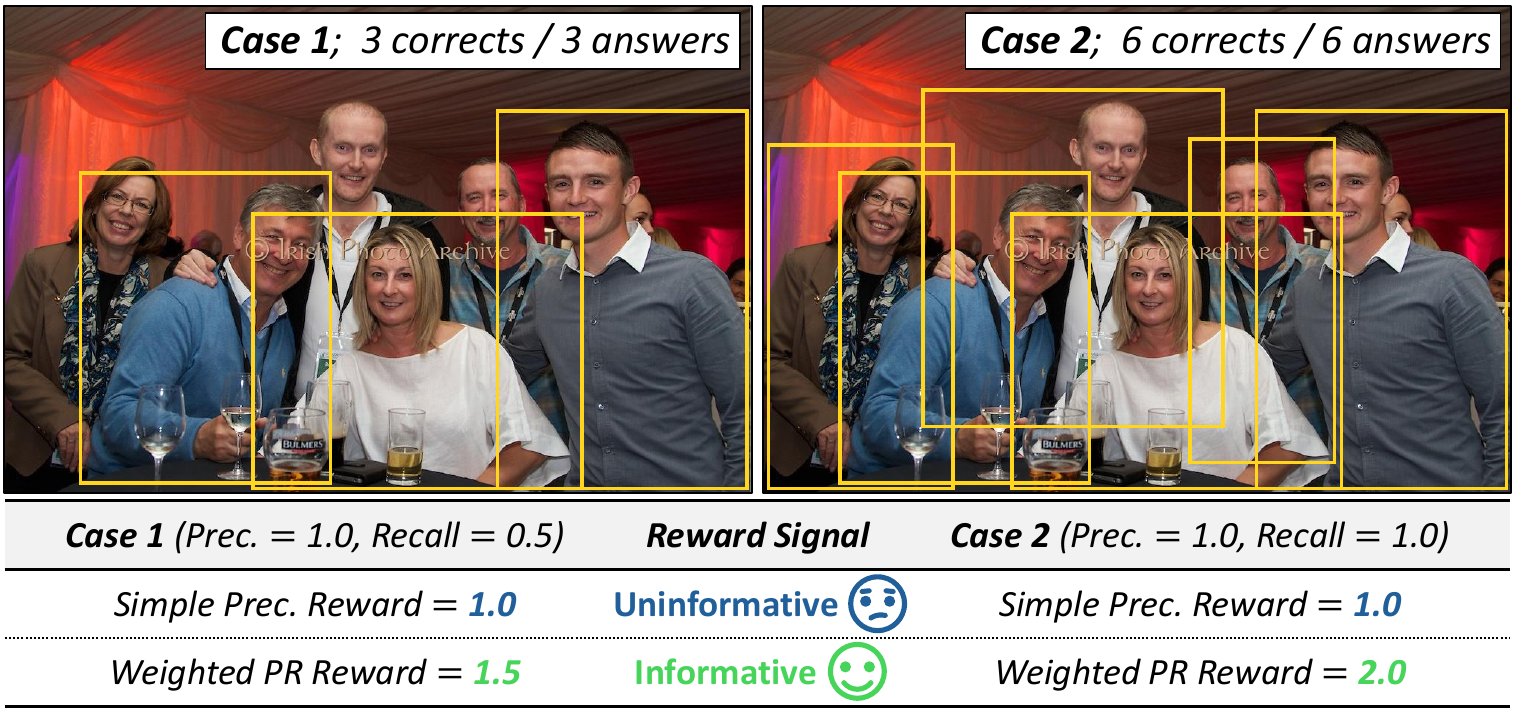}}
    \end{minipage}
	\caption{Orange boxes are answers. While a simple precision reward provides uninformative signal giving same reward value for both cases, our weighted precision-recall reward gives informative signal distinguishing two cases.}
	\label{fig6}
\end{figure}

\section{Discussion}
\noindent
\textbf{Intuition of Precision-Recall Accuracy Reward }As described in Section~\ref{pr-reward}, we design a weighted precision-recall reward function, which assigns high rewards when both precision and recall are high. In usual precision-recall curves (which slope downward to the right), there exists an inherent trade-off: high recall is often along with low precision. A simple precision-based reward fails to consider this trade-off. For instance, as shown in Fig.~\ref{fig6}, it would assign equal rewards for different cases, even though the right prediction is more desirable. It means that there is no any guidance to minimize false negatives and to cover target instances as many as possible. On the other hand, our weighted precision-recall reward distinguishes two cases, encouraging the model to answer in a desirable way covering many target instances.

\noindent
\textbf{Applicability and Purpose of DIP-R1 }To verify the general applicability of our framework across diverse vision tasks, we further evaluate DIP-R1 on COCO \cite{coco}--a standard benchmark for general object detection--and RefCOCO \cite{refcoco}--a benchmark for visual grounding. The experimental results presented in TABLE~\ref{tab5} demonstrate the effectiveness of DIP-R1 in extending to other vision-language tasks. Moreover, such analyses confirm that the goal of DIP-R1 is to enhance the fine-grained visual perception capabilities of MLLMs, particularly under visually complex conditions. For instance, our framework can be beneficial for interactive applications such as crowd monitoring and public safety, where it enables identifying individuals performing specific actions or carrying particular objects in crowded environments like airports.

\begin{figure}[t!]
    \begin{minipage}[b]{0.99\linewidth}
        \centering
        \centerline{\includegraphics[width=\linewidth]{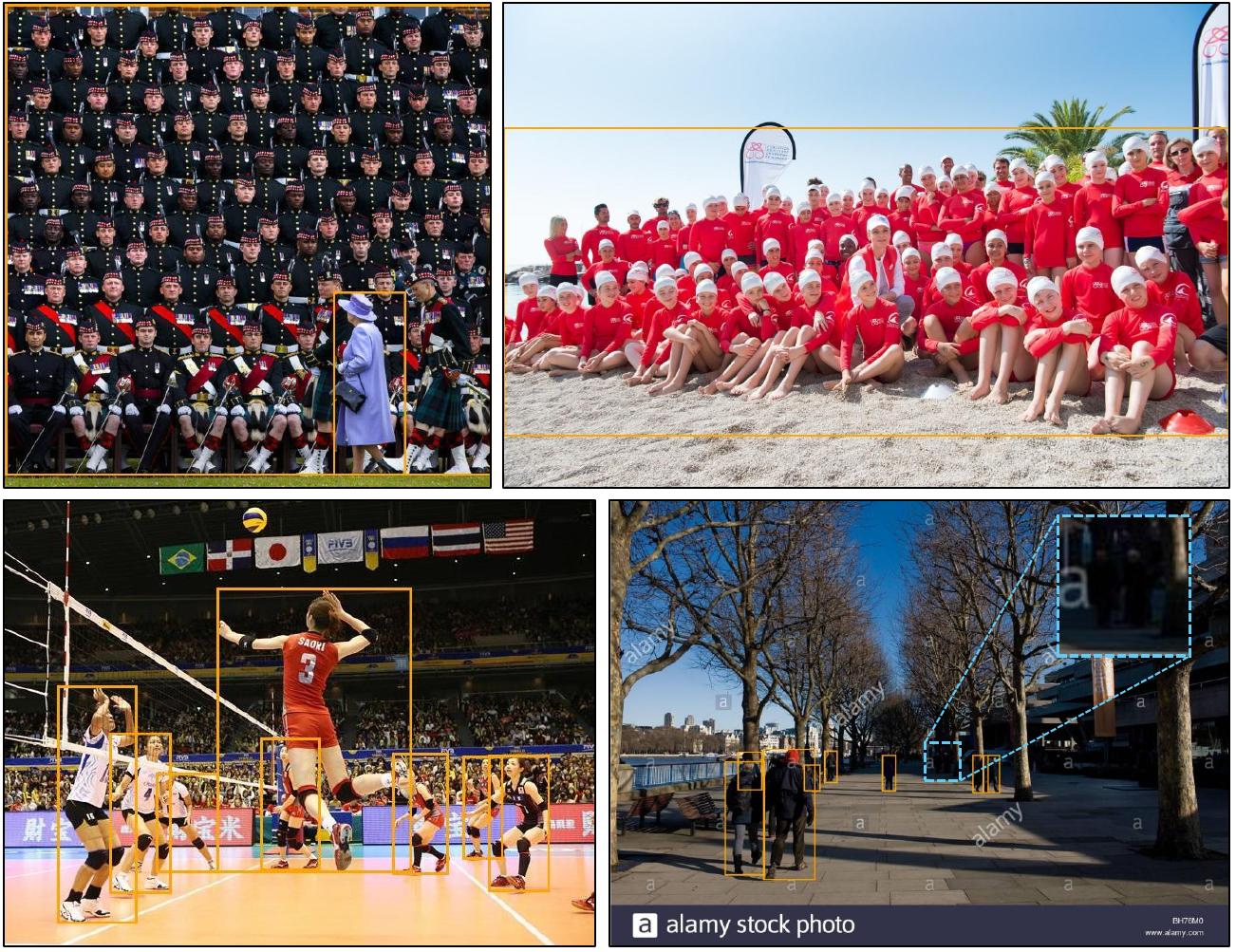}}
    \end{minipage}
	\caption{Failure cases of the proposed framework. The framework still struggles to perceive instances (1) with highly repetitive patterns, (2) which are too small or blurred, and (3) under low-light conditions. The yellow box indicates the perception result, while the blue dotted box is the zoomed-in region of low-light instances.}
	\label{fig7}
\end{figure}

\begin{figure*}[!h]
    \centering
    \includegraphics[width=0.98\textwidth]{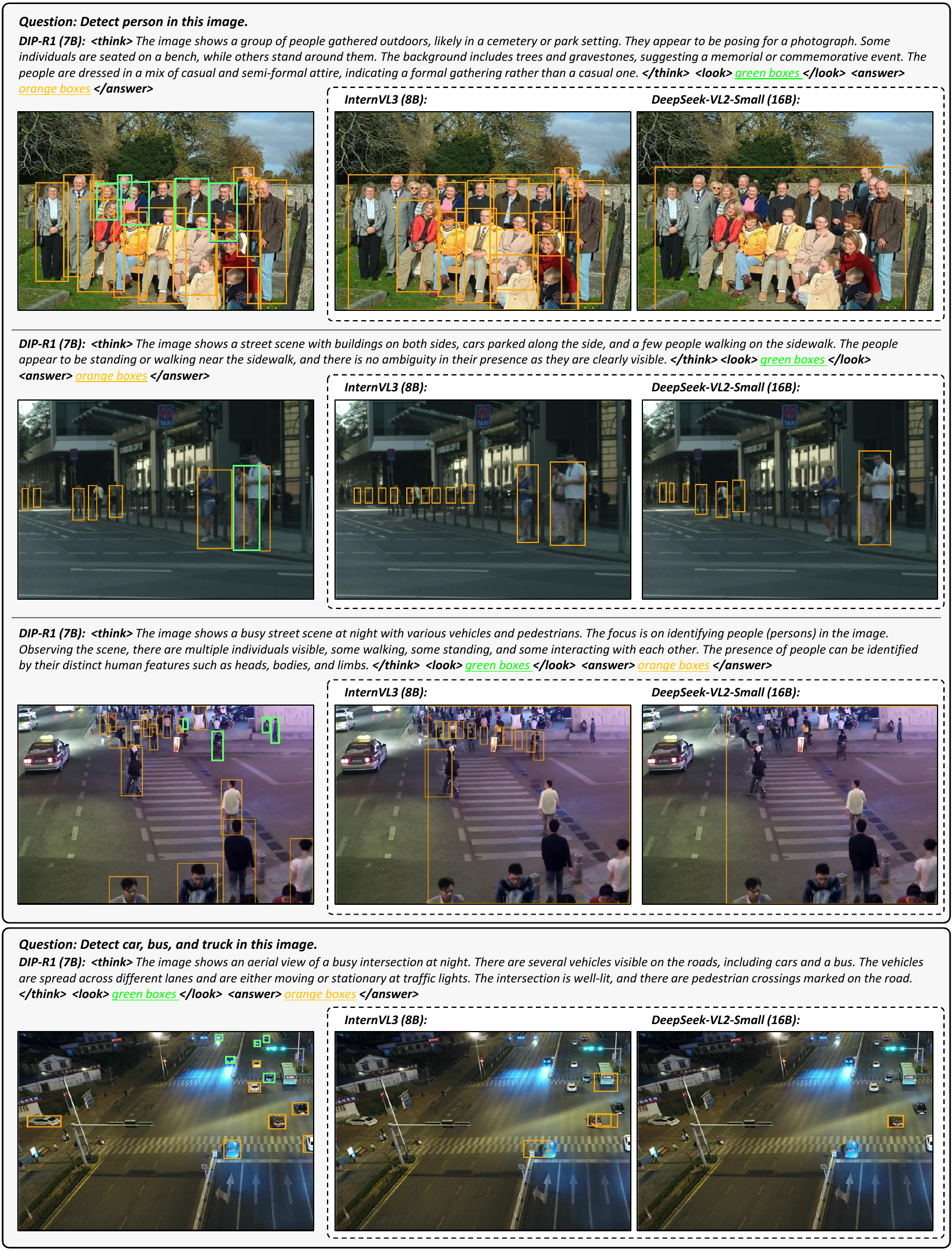}
    \vspace{-0.1cm}
	\caption{Additional qualitative visualization results. The results the response results of DIP-R1 including its reasoning descriptions, uncertain regions (\textit{green boxes}) captured in the observing process, and the prediction results (\textit{orange boxes}).}
	\label{fig8}
\end{figure*}

\noindent
\textbf{Failure Cases }Even though the proposed DIP-R1 obtains noticeable enhancements in visual perception of MLLM under complex scenarios, it still encounters difficulties in recognizing under extremely challenging conditions. Fig.~\ref{fig7} illustrates representative failure cases of our framework. As shown in the top-left and right examples, DIP-R1 struggles to distinguish individual instances when highly repetitive patterns are present. In the bottom-left and right examples, it also fails to accurately perceive small, blurred, or low-light instances that are barely visible.

\begin{table}[t!]
    \centering
    \caption{The applicability on general object detection COCO validation set (AP$^{val}_{50}$) and visual grounding RefCOCO validation set (Acc).}
    \renewcommand{\arraystretch}{1.1}
    \renewcommand{\tabcolsep}{5.0mm}
    \resizebox{0.999\linewidth}{!}
    {
        \renewcommand{\arraystretch}{1.3}
        \begin{tabular}{l|cc}
            \toprule[1.3pt]
            \bf Method & \bf COCO & \bf RefCOCO \\ \midrule
            DeepSeek-VL2-Tiny (3B) \cite{deepseek-vl2} & -- & 84.7 \\
            Qwen2.5-VL-Instruct (3B) \cite{qwen2.5-vl} & 59.3 & 85.5 \\
            Shikra (7B) \cite{shikra} & 40.3 & 87.0 \\ \midrule
            DIP-R1 (3B) & \bf 66.2 & \bf 87.5 \\
            \bottomrule[1.3pt]
        \end{tabular}
    }
    \label{tab5}
\end{table}

\noindent
\textbf{Limitation }In this work, we explore the effectiveness RL and validity of our designed reward modelings, such as variance-guided look reward, and we focus on evaluating our DIP-R1 framework in the context of fine-grained object detection using extensive in-domain and out-of-domain datasets. However, it can be extended to more diverse vision-language tasks, so that we plan to broaden our framework to other tasks, such as image captioning with multiple region-level groundings, which would further demonstrates its applicability across a wider range of vision-language problems.

\section{Conclusion}
In this work, we proposed DIP-R1, a novel reinforcement learning (RL) framework designed to enhance the fine-grained visual perception capabilities of Multimodal Large Language Models (MLLMs) in complex real-world environments. We employed three simple yet effective rule-based reward functions for step-by-step processes, \textit{reasoning}, \textit{observing}, and \textit{decision-making}. Due to conventional format reward, reasoning process includes scene-level understanding process. Moreover, we design a variance-guided reward function for the model to look through uncertain regions in the observing process in instance level. Then a weighted precision-recall reward helps generate more accurate and robust answer within complex real-world scenes. Extensive experiments over four challenging object detection tasks demonstrate the effectiveness of DIP-R1 framework showing remarkable performance improvement. These results corroborate the potential of RL to boost deep visual perception ability of MLLMs in real-world scenarios.

\bibliographystyle{IEEEtran}
\bibliography{IEEEabrv,main}

\end{document}